\documentclass[letterpaper, 10 pt, conference]{ieeeconf}  

\IEEEoverridecommandlockouts                              

\usepackage{graphicx}
\usepackage{comment}
\usepackage{amsmath,amssymb} 
\usepackage{color}

\usepackage{hyperref}       
\usepackage{url}            
\usepackage{booktabs}       
\usepackage{amsfonts}       
\usepackage{nicefrac}       
\usepackage{microtype}      

\usepackage{times}
\usepackage{epsfig}
\usepackage{graphicx}
\usepackage{amsmath}
\usepackage{amssymb}
\usepackage{gensymb}
\usepackage{dsfont}

\usepackage{units}                                                                                                                        
\usepackage{subcaption}
\usepackage{multirow}
\usepackage{bm}
\usepackage{mydefs}
\usepackage{bbm}
\usepackage{pifont}
\usepackage{xspace}
\usepackage{float}

\usepackage{graphicx,calc} 
\usepackage{color}    
\usepackage{ifthen}
\usepackage{paralist}
\usepackage{times}
\usepackage{longtable}
\usepackage{colortbl}
\usepackage{nomencl}
\usepackage{subcaption}
\usepackage{bm}
\usepackage[maxfloats=256]{morefloats}
\usepackage{balance}

\newcommand{\ie}{\textit{i.e.}\xspace}
\newcommand{\eg}{\textit{e.g.}\xspace}
\newcommand{\etc}{\textit{etc}\xspace}
\newcommand{\etal}{\textit{et al.}\xspace}

\newcommand{\out}[1]{}

\title{\LARGE \bf
Identifying Driver Interactions via Conditional Behavior Prediction
}

\author{Ekaterina Tolstaya$^{1}$, Reza Mahjourian$^{2}$, Carlton Downey$^{2}$,\\ Balakrishnan Varadarajan$^2$, Benjamin Sapp$^2$, Dragomir Anguelov$^{2}$  
\thanks{$^{1}$ General Robotics Automation Sensing and
Perception (GRASP) Laboratory at the University of Pennsylvania,  {\tt\small eig@seas.upenn.edu}}%
\thanks{$^{2}$ Waymo, {\tt\small rezama@waymo.com}}%
\thanks{The first author acknowledges support from the NSF Graduate Research Fellowships Program.}
}

\begin{document}

\maketitle


\begin{abstract}
Interactive driving scenarios, such as lane changes, merges and unprotected turns, are some of the most challenging situations for autonomous driving. Planning in interactive scenarios requires accurately modeling the reactions of other agents to different future actions of the ego agent. We develop end-to-end models for conditional behavior prediction (CBP) that take as an input a query future trajectory for an ego-agent, and predict distributions over future trajectories for other agents conditioned on the query.  Leveraging such a model, we develop a general-purpose agent interactivity score derived from probabilistic first principles.
The interactivity score allows us to find interesting interactive scenarios for training and evaluating behavior prediction models. We further demonstrate that the proposed score is effective for agent prioritization under computational budget constraints.
\end{abstract}

\section{Introduction}

Behavior prediction is a core component of real-world systems involving human-robot interaction.
This task is particularly challenging due to the high degree of uncertainty in the future---the intent of human actors is unobserved, and 
multiple interacting agents may continually influence one another.

\begin{figure}[t]
      \centering
      \includegraphics[width=0.45\textwidth]{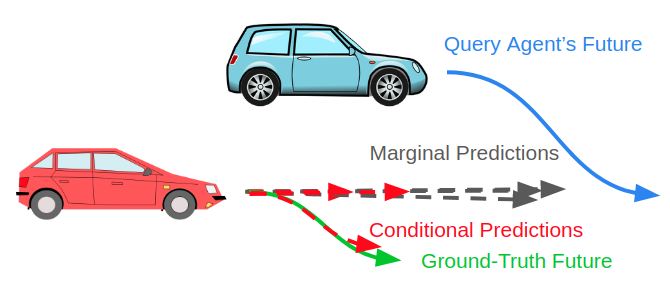}
      \caption{A conditional behavior prediction model describes how one agent's predicted future trajectory can shift due to the actions of other agents.} \vspace{-0.5cm}
      \label{cartoon2}
  \end{figure}

We are particularly interested in the high-impact application of Autonomous Vehicles (AV), in which a robot may wish to pose behavior prediction queries of the form ``If I take action $X$, what will agent $B$ do?'', as shown in Figure~\ref{cartoon2}.   We assert that this type of {\em conditional inference} is important and fundamental for making planning decisions in an interactive driving environment.  In this paper, we focus on probabilistic models of future behavior that can condition on possible future action sequences (\ie, trajectories) of other agents.  We call this task {\em conditional behavior prediction} (CBP).

In the literature, there are a family of behavior prediction models for which the conditioning capability comes naturally: those that employ step-wise, iterative sampling (``roll-outs'') for multiple agents in a scene, \eg~\cite{sociallstm, tang_multifuture,precog_Rhinehart_2019_ICCV, schmerling2018multimodal}. In such models, it is possible to control the action sequences for a subset of agents, so that the roll-out of others will take them into account. While flexible, these sample-based models have significant disadvantages for real-world applications: sample-based inference is risky to employ in a safety-critical system, iterative errors can compound~\cite{ross2011dagger}, it is difficult to control sample diversity~\cite{rhinehart2018r2p2, kitani_diverse_forecasting_dpps}, and attempting to jointly model all agents is often intractable computationally.  Some past work has focused on tightly-coupled robot-human interaction in limited driving game environments: ~\cite{schmerling2018multimodal} iteratively conditions on generated human actions in a CVAE framework;~\cite{sadigh2016planning} formulates the interaction problem as a 2-player game with human reward learned via inverse reinforcement learning.

As an alternative to sample-based models, there is a long line of work on single-shot, {\em passive} behavior prediction~\cite{chai2019multipath,DESIRE,neural_motion_planner_zeng2019,casas2018intentnet,SocialGAN,gao2020vectornet, chang2019argoverse}.  These models are compelling due to tractability and practical parametric output distributions, and have become the popular choice in AV systems and associated benchmarks~\cite{chang2019argoverse}.  However, these models ignore the fact that the AV ego-agent will take actions in the future, which may cause a critical reaction by another agent.  Using such models makes decision-making challenging: because the models do not condition on any explicit ego actions, they must implicitly account for all possible ego-actions (or ignore interactions altogether).  In practice,  interaction modeling has been handled via aggregating neighboring agents' observed states via max-pooling, transformer layers, or graph neural network architectures~\cite{sociallstm, mercat2020multi, casas2020spagnn,  mangalam2020_journey_pecnet}.

In this paper, we propose a single-shot, {\em conditional} behavior prediction model. Our CBP model is a powerful, end-to-end trained deep neural network, which takes into account static and dynamic scene elements---road lanes, agent state histories (vehicle, pedestrian and cyclist), traffic light information, \etc.  From these inputs, we predict a diverse set of future outcomes, represented as Gaussian Mixture distributions, where each mixture component corresponds to a future state sequence (\ie, a trajectory with uncertainty).  We train these models to be capable of conditional inference by selectively adding future trajectory information for some agents as additional inputs to the model.  We use large datasets of logged driving data and train models via maximum likelihood to output either conditional or passive (marginal) predictions for any subset of agents in a scene.  The recently proposed WIMP model~\cite{wimp2020} is also a single-shot conditional inference model; ours differs in that we condition on generic trajectories for any subset of agents in a fully probabilistic framework.

The notion of interactivity is a key concept for this problem, and a key contribution of this paper is to formalize the notion and obtain a simple and practical {\em interactivity score}.
Now that we are equipped with a probabilistic model for conditional future distributions, we can quantify a notion of interactivity as follows.  We quantify the {\em degree of influence} one agent has on another as the KL-divergence between (a) the agent's future distribution conditioned on the other's future and (b) its marginal distribution.  We then take an expectation over all possible conditioned futures for the other agent to get a final interactivity score.  This results in a simple, agent-symmetric computation in the form of {\em mutual information} between the two agent's futures.  In contrast, past work have hand-designed models of surprise or discomfort for motion planning~\cite{pandey2010framework, scandolo2011anthropomorphic, sisbot2007human,refaat2019agent}. Entropy and mutual information have been previously used in AV applications as a measure of uncertainty to predict collisions~\cite{michelmore2018evaluating}.



In real-world driving, the interactivity score can be used to anticipate driver interactions.  When processing data offline, we demonstrate the use of the interactivity score for mining interactive scenarios that are potentially unsafe, since the target agent's expectations are being violated. 
Furthermore, we demonstrate the benefits of the interactivity score for prioritizing agents for behavior prediction and planning. In contrast to previous work that built a special-purpose model trained directly for the task of prioritization~\cite{refaat2019agent}, which was derived from an implicitly-defined side-channel output of a blackbox planner, we provide a formulation that is independent of a specific planner definition and consequently, more generally applicable.

Our contributions can be summarized as follows: (1) We provide a novel, principled information-theoretic definition of interactivity, which applies to any multi-agent interaction application, (2) we develop a first-of-its-kind, single-shot, deep neural network for probabilistic conditional behavior prediction and (3) we show our interactivity score improves state-of-the-art model performance in several settings.

\section{Defining Agent Interactivity}

 We define an agent {\em trajectory} $S$ as a fixed-length, time-discretized sequence of agent states up to a finite time horizon. All quantities in this work consider a pair of agents $A$ and $B$.  Without loss of generality, we consider $A$ to be the {\em query} agent whose plan for the future can potentially affect $B$, the {\em target} agent. The future trajectories of $A$ and $B$ are random variables $S^A$ and $S^B$. The {\em marginal} probability of a particular realization of agent $B$'s trajectory $\bms^B$ is given by $p(S^B=\bms^B)$, also indicated by the shorthand $p(\bms^B)$. The {\em conditional} distribution of agent $B$'s future trajectory given a realization of agent $A$'s trajectory $\bms^A$ is given by $p(S^B = \bms^B | S^A = \bms^A)$, indicated by the shorthand $p(\bms^B | \bms^A)$.

Even in highly interactive scenarios, agents may behave as expected by other agents and not exert any influence on one another.  A define a {\em surprising interaction} as one in which the target agent experiences a change in their behavior due to the query agent's observed trajectory. 
When we have access to ground-truth future trajectories, we can quantify interactions by estimating the change in log likelihood of the target's ground-truth future $\bms^B$:
\begin{equation} \label{eq:diff_log_likelihood}
   \Delta_{\text{LL}} 
   := \log  p(\bms^B | \bms^A) - \log  p(\bms^B) %
\end{equation}
A large change in the log-likelihood indicates a situation in which the likelihood of the target agent's trajectory changes significantly as a result of the query agent's action.  If the target's trajectory $\bms^B$ becomes more likely given the query agent's trajectory $\bms^A$, then $\Delta_{\text{LL}}$ will be positive. If it becomes less likely, then  $\Delta_{\text{LL}}$ will be negative.  If there is no change, then $\Delta_{\text{LL}}$ will be zero.

A query agent may need to estimate the impact of a planned future trajectory $\bms^A$ on the target agent $B$.  Since we don't have access to the ground-truth future for the target agent, we can quantify the potential for a surprising interaction by estimating the shift in the distribution of the target agent's trajectory.  More specifically, we use the KL-divergence between the conditional and marginal distributions for the target's predicted future trajectory $S^B$ to quantify the degree of influence exerted on $B$ by a a trajectory $\bms^A$:
\begin{equation} \label{eq:kl_div_def}
 D_{\text{KL}} \left \lbrack p(S^B | \bms^A) \middle\| p(S^B) \right \rbrack = \!\! \int_{\bms^B} p(\bms^B | \bms^A) \log  \frac{p(\bms^B | \bms^A)}{ p(\bms^B)}
\end{equation}
For example, in Fig. \ref{cartoon2}, if the query agent decides to change lanes in front of the target agent, the target agent will have to slow down.  In this case, the KL-divergence will reflect a significant change in the target agent's expected behavior as a result of the query agent's planned lane change.
In the absence of a particular plan for the query or target agent, we can consider the set of all possible actions for the query agent and compute the expectation of the degree of influence over all those possible actions.  This expectation is defined as the \emph{mutual information} between the two agents' future trajectories $S^A$ and $S^B$, and is computed as:
\begin{equation} \label{eq:mutual_info}
 I(S^A, S^B) = \int_{\bms^A} p(\bms^A)   D_{\text{KL}} \left \lbrack p(S^B |  \bms^A) \middle\| p(S^B) \right \rbrack  
\end{equation}
Mutual information expresses the dependence between two random variables.  It is non-negative, $I(S^A, S^B) \geq 0$, and symmetric, $I(S^A, S^B) = I(S^B, S^A)$ \cite{shannon1948mathematical}.  We use this quantity as the {\em interactivity score} between agents $A$ and $B$. For example, if the target agent is driving closely behind the query agent, we expect their interactivity score to be high because the target agent is likely to respond immediately to any actions, such as deceleration or acceleration, from the query agent.

\section{Method}

In the previous section, we developed a measure of interactivity between a pair of agents.  In this section, we discuss training a conditional behavior prediction model that can estimate the distributions $p(\bms^B)$ and $p(\bms^B | \bms^A)$.  We discuss the internals of this model and losses.  We then discuss the process for computing the interactivity score by sampling from the predicted distributions.


Let $\bmx$ denote observations from the scene, including past trajectories of all agents, and context information such as lane semantics. Let $t$ denote a discrete time step, and let $s_t$ denote the state of an agent at time $t$. The realization of the future trajectory $\bms = \lbrace s_1, \ldots, s_T \rbrace$ is a sequence of states for $t \in \{1, \dots, T\}$, a fixed horizon.

 A CBP model predicts $p(S^B | S^A \!\! = \bms^A, \bmx)$, the distribution of future trajectories for $B$ conditioned on $\bms^A$.
The CBP model receives as input a realization of the future trajectory of the query agent, $\bms^A = \lbrack s_1^A, \ldots, s_T^A \rbrack$, which we refer to as agent $A$'s \emph{plan}, or the \emph{conditional query}.
Following the approach of MultiPath \cite{chai2019multipath}, the model predicts a set of $K$ trajectories for agent $B$, $\bmmu^B = \lbrace \bmmu^{Bk} \rbrace^K_{k=1}$, where each trajectory is a sequence of states $\bmmu^{Bk} = \lbrace \bmmu_1^{Bk}, \ldots, \bmmu_T^{Bk} \rbrace$, capturing $K$ potentially-different intents for agent $B$.  The model predicts uncertainty over the $K$ intents as a softmax distribution $\pi^{Bk}(\bmx, \bms^A)$.  The model also predicts Gaussian uncertainty over the positions of the trajectory waypoints as:
\begin{equation} \label{eq:conditional_likelihood}
    \phi^{Bk}(s^B_t | \bmx, \bms^A) = \mathcal{N} \left( s^B_t | \bmmu_t^{Bk}(\bmx, \bms^A), \Sigma_t^{Bk}(\bmx, \bms^A)\right).
\end{equation}
This yields the full conditional distribution $p(\hat{S}^B | \bms^A, \bmx)$ as a Gaussian Mixture Model (GMM) with mixture weights fixed over all time steps of the same trajectory:
\begin{equation} \label{eq:cbp_gmm}
    p(\hat{S}^B \!= \bms^B|\bmx, \bms^A) = \!\!\sum_{k=1}^K \pi^{Bk}\!(\bmx, \bms^A) \!\prod_{t=1}^T \phi^{Bk}(s_t^{B} | \bmx, \bms^A).
\end{equation}
The Gaussian parameters $\bmmu_t^{Bk}$ and $\Sigma_t^{Bk}$ are directly predicted by a deep neural network (DNN).  The softmax distribution is computed as $\pi^{Bk}(\bmx, \bms^A) = \frac{\exp f_{k}^B(\bmx, \bms^A)}{\sum_i \exp f_i^B(\bmx, \bms^A)}$, where $f_k^B$ are logit values also output by the DNN.

The computation of the interactivity score also requires the estimation of marginal distributions, $p(S^B|\bmx)$, which are not conditioned on any future plan for $A$.
We train a single model which can produce both marginal and conditional predictions, in order to have comparable quantities without any uncertainty due to model variance. 
Marginal predictions, $p(\tilde{S}^B | \bmx)$, are provided by turning off inputs from the conditional query encoder in the model.  We adopt the shorthands $\pi^{Bk}(\bmx, \emptyset), \phi^{Bk}(\bmx, \emptyset)$ to describe this operation, which gives us the marginal distribution as
\begin{equation} \label{eq:bp_gmm}
    p(\tilde{S}^B = \bms^B|\bmx) = \sum_{k=1}^K \pi^{Bk}(\bmx, \emptyset) \prod_{t=1}^T \phi^{Bk}(s_t^{B} | \bmx, \emptyset).
\end{equation}

Given the conditional and marginal predictions of the CBP model, we can now compute the mutual information.  Directly computing the mutual information between the future states of two agents via Eq.~\eqref{eq:mutual_info} is intractable between the GMM distributions (Eq.~\eqref{eq:bp_gmm}). 
We estimate the outer expectation via importance sampling. Rather than sampling $N$ samples from the marginal distribution, we will use the most likely 6 modes of the marginal distribution’s GMM in Eq.~\eqref{eq:bp_gmm} as in standard motion forecasting metrics \cite{chang2019argoverse}, with  $\bms_k^{A} \in \lbrace \bmmu^{kA}(\bmx)\rbrace_{k=1}^6$: 
\begin{equation}
 I(S^A\!\!, S^B | \bmx) \! \approx \!\! \frac{1}{M} \!\!\sum_{k=1}^6 \! \sum_m  p(\bms_k^{A} | \bmx) \log \frac{p(\hat{S}^B \!\! = \! \bms_m^B | \bms_k^{A}, \bmx)}{p(\tilde{S}^B  \! = \bms_m^B | \bmx)} 
\end{equation}
where the marginal and conditional probabilities are evaluated via Eqs. \eqref{eq:cbp_gmm} and \eqref{eq:bp_gmm}.
The use of other more efficient approaches for estimating KL divergence are left to future work \cite{hershey2007approximating}. 


To train the model for conditional prediction, we set the conditional query/plan input to agent $A$'s ground-truth future trajectory from the training dataset.  
We learn to predict the distribution parameters  $f_{k}^B(\bmx, \bms^A)$, $\bmmu_t^{Bk}(\bmx, \bms^A)$, and  $\Sigma_t^{Bk}(\bmx, \bms^A)$ via supervised learning with the negative log-likelihood loss,
\begin{equation} \label{eq:loss}
\begin{split}
    \mathcal{L}(\theta) = \ \sum_{m=1}^M \sum_{k=1}^K \mathds{1} (k = k_m^B) \Big\lbrack \log \pi^{Bk} (\bmx_m, \bms^A_m; \theta) \\+  \sum_{t=1}^T \log \mathcal{N} (s_t^{Bk} | \bmmu_t^{Bk}, \Sigma_t^{Bk};  \bmx_m, \bms^A_m; \theta) \Big\rbrack,
\end{split}
\end{equation}  
where $k_m$ is the index of the mode of the distribution that has the closest endpoint to the given ground-truth trajectory, $k_m^B =  \argmin_k \sum_{t=1}^T \lVert s_t^{Bk} - \bmmu_t^{Bk}\rVert_2$.
%

Above, we describe how to produce predictions for a single agent $B$.  However, for increased efficiency, our model produces predictions for multiple agents in parallel.  To encourage the model to maintain the fundamental physical property that agents cannot occupy the same future location in space-time, we include an additional loss function:
\begin{equation}
    \mathcal{L}_\mathcal{O}(\theta) = \! \sum_i \sum_j \pi^{Ai} \pi^{Bj} \max_t \exp( -\lVert\bmmu^{Ai}_t \!-\bmmu^{Bj}_t\rVert_2^2 / \alpha),
\end{equation}
where $\lbrace  (\pi^{Ai}, \bmmu^{Ai})\rbrace_{i=1}^K$ and $\lbrace  (\pi^{Bj}, \bmmu^{Bj})\rbrace_{j=1}^K$ are the modes and probabilities of the future trajectory distributions for agents $A$ and $B$.

\section{Experiments}
\subsection{Data}

We collected a large, in-house dataset of real-world driving from urban and suburban environments, using a vehicle equipped with an industry-grade sensor and perception stack, which provides us with tracked objects.  In total, the training set has 1.9 billion vehicle agents that we learn to model, from 19 million unique scenarios, comprising 18 years of continuous driving data.  The models receive 2 seconds of history and predict 15 seconds of future behavior for all agents in the scene, including the AV.  The state of the agents are recorded at 5 Hz.  Features describing the past states of the agents include $(x, y, z)$ position, velocity vector, acceleration vector, orientation $\theta$, and angular velocity. There are also binary attributes indicating whether the vehicle is signaling to turn left or right, and whether it is parked.  The lane markings and boundaries are represented by 500 points sampled around the current location of each predicted vehicle to balance memory requirements.

At training time, we select one agent uniformly at random from the vehicles in the scene to be the query agent.  For 95\% of the samples, the query agent's future ground-truth is fed to the model as the conditional query input. For the other 5\%, no conditional query is provided, leading to marginal behavior prediction, with the split chosen through cross-validation. 

For every scene, the model predicts future behaviors for up to the 20 closest vehicles.  Other agents (vehicles, pedestrians, and cyclists) are still used in the agent state feature encoder, but the model doesn't predict futures for them.  
We observed that prediction performance beyond 20 agents degrades rapidly due to sensor limitations. 

\subsection{Model Architecture}

  \begin{figure}[t]
      \centering
      \includegraphics[width=0.5\textwidth]{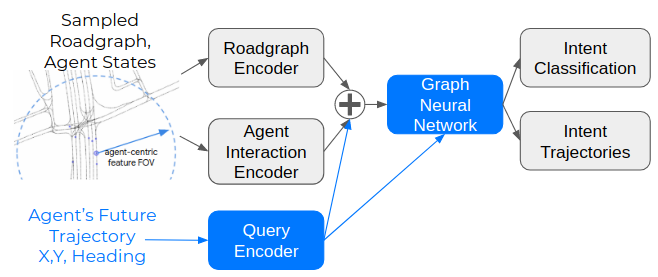}
      \caption{The architecture of the conditional behavior prediction model.}
      \label{architecture} \vspace{-0.5cm}
  \end{figure}

The architecture is composed of an input encoder stage, a trajectory decoder stage, and a GNN-based trajectory refinement stage, shown in Fig. \ref{architecture}. 
The encoder stage is composed of a road lane encoder which uses an architecture similar to VectorNet~\cite{gao2020vectornet}, and  
a track history encoder which uses a 64-dimensional LSTM applied to 5 time steps of past state observations comprising 1 second of history.
The result of the above two encoders are concatenated and passed into a decoder which outputs a sequence of $(x,y)$ points via predicted polynomial coefficients for $K=287$ trajectory modes \cite{chai2019multipath}. In our experiments, we use a tenth-degree polynomial. The resulting trajectories are further refined using a GNN~\cite{Battaglia2018GNNs, casas2020spagnn}. The GNN uses an attention-based aggregation function that combines relative agent positions as edge features to form messages passed to each node \cite{vaswani2017attention}.  We apply one message update, which passes trajectory information between neighboring agents, and then re-apply trajectory decoding.  This process can refine the agents' trajectory distributions with awareness of their neighbors' distributions.
Further details of this state-of-the-art model architecture are currently under anonymous review.

\section{Results}
\subsection{Metrics}

Given a labeled example $(\bmx, \bms^A, \bms^B)$, the weighted Average Distance Error (wADE) over the most likely 6 modes of the conditional prediction of agent $B$'s future trajectory given the query agent $A$'s future trajectory is:
\begin{equation}
\text{wADE}^6_\text{CBP}(B) = \frac{1}{T} \sum_t \sum_{k=1}^6 \pi^{Bk}( \bms^A, \bmx) \lVert s_t^B - \bmmu^{Bk}_t \rVert_2,
\end{equation}
where $\bmmu^{Bk}_t = \bmmu_t^{Bk}(\bms^A, \bmx)$ is the $k$th mode for the predicted position of agent $B$ at time $t$ with its respective probability of 
$\pi^{Bk}( \bms^A, \bmx)$. Likewise, we can compute the $\text{wADE}_{\text{BP}}$ metric using $\bmmu^{Bk}_t =\bmmu^{Bk}_t(\bmx, \emptyset)$ and $\pi^{Bk}(\bmx, \emptyset)$.
Computing their difference:
$
    \Delta_{\text{wADE}} (B) = \text{wADE}_{\text{BP}} (B)  - \text{wADE}_{\text{CBP}} (B)
$
quantifies the reduction in $B$'s error due to conditioning on $A$.
Another established metric for behavior prediction is the minimum Average Distance Error (minADE), defined for conditional models over the most likely 6 modes as:
\begin{equation}
    \text{minADE}^6_\text{CBP}(B) = \min_{1 \leq k \leq 6}  \frac{1}{T} \sum_t  \lVert s_t^B - \bmmu^{Bk}_t(\bms^A, \bmx) \rVert_2.
\end{equation}

To obtain a low minADE value, the model needs to accurately predict the ground-truth future as one of its predicted intents.  On the other hand, the wADE metric is more suitable for evaluating multi-modal distributions and can reflect shifts in distribution of intent probabilities.  Therefore, we use wADE as the main metric in the following results. Furthermore, $\Delta_{\text{wADE}}$ is closely related to the definition of $\Delta_{\text{LL}}$ in Eq.~\eqref{eq:diff_log_likelihood}, but since it is weighted by the distance error, it is less sensitive to prediction errors for nearly-stationary vehicles.

\subsection{Conditional Behavior Prediction} Comparing accuracy between marginal and conditional predictions from the trained model shows a 10\% improvement for conditional predictions, as seen in Table \ref{tab:wADE}.  This is clear confirmation that our model is leveraging future information to improve predictive power, as expected. The early fusion conditional encoder receives the conditional query at an earlier stage in the model, whereas the late fusion setup feeds the query to the GNN only at the final prediction stage.  As the results show, the early-fusion variant significantly outperforms late fusion.

\subsection{Evaluation on Argoverse}

Our model is competitive with state of the art on the popular Argoverse benchmark dataset \cite{chang2019argoverse}.  On the validation dataset, we achieve a minADE$_6$ of 0.7488, which is near state of the art in recent work: 0.71 by Liang \etal \cite{liang2020learning}, 0.728 by TNT \cite{zhao2020tnt}, and 0.75 by WIMP \cite{wimp2020}. By conditioning on the sensor vehicle, the CBP model reduces minADE by 0.8\% to 0.7409, consistent with our more exhaustive studies on the internal dataset. 

\begin{table}[!tbp]
\caption{\small Comparison of CBP models on an evaluation dataset containing over 8 million agent pairs. Metrics are computed and averaged over all (query agent, target agent) pairs possible in every scene.  The mean error is computed only over predictions for the target agent and does not include predictions for the query agent.  The standard error of the mean is also reported.}
\label{tab:wADE}
\centering
\begin{tabular}{lcc}
\toprule
\multirow{1}{*}{Method} & $\text{wADE}^6_{\text{CBP}}(B) \downarrow$  & $\text{minADE}^6_{\text{CBP}}(B) \downarrow$    \tabularnewline
\midrule
Non-conditional         & 3.486 $\pm$ 0.0017   & 1.207 $\pm$ 0.00062
 \tabularnewline
Early fusion (encoder)          & 3.142  $\pm$ 0.0016       & 1.170 $\pm$ 0.00061  \tabularnewline
Late fusion (GNN)              & 3.469 $\pm$ 0.0017         & 1.209 $\pm$ 0.00063  \tabularnewline
Early and late fusion    & 3.160 $\pm$ 0.0016         & 1.172 $\pm$ 0.00067  \tabularnewline
\bottomrule
\vspace{-0.2cm}
\end{tabular}
\vspace{-0.5cm}
\end{table}

\subsection{Distribution of Interactivity Scores}

\begin{figure}[h]
      \centering
      \includegraphics[width=0.35\textwidth]{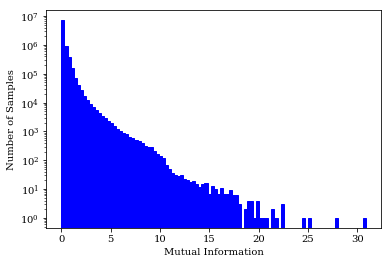}
      \caption{Histogram of interactivity score (mutual information) between 8,919,306 pairs of agents in the validation dataset. }
      \vspace{-0.5cm}
      \label{mi_histogram}
  \end{figure}

Figure \ref{mi_histogram} shows the histogram of interactivity scores between all agent pairs in the evaluation dataset.  
The incidence of interactions in most datasets are rare, so the interactivity score may be a good tool to automatically mine a dataset for interactive examples.

\subsection{Interactivity Score Predicts Surprise} The mutual information score allows us to discover scenarios with a potential for surprising interactions, where the ground-truth future of the query agent causes a target agent to change its behavior.
Using the ground-truth future trajectories of agents $\bms^A$ and $\bms^B$, we can quantify how query agent $A$ affected target agent $B$ in reality by comparing the prediction error between the conditional and marginal (non-conditional) models.
A large, positive $\Delta_{\mbox{wADE}}$ indicates that providing the query agent’s future significantly improves the prediction accuracy for the target agent.

\begin{figure}[t] 
      \includegraphics[width=0.48\textwidth]{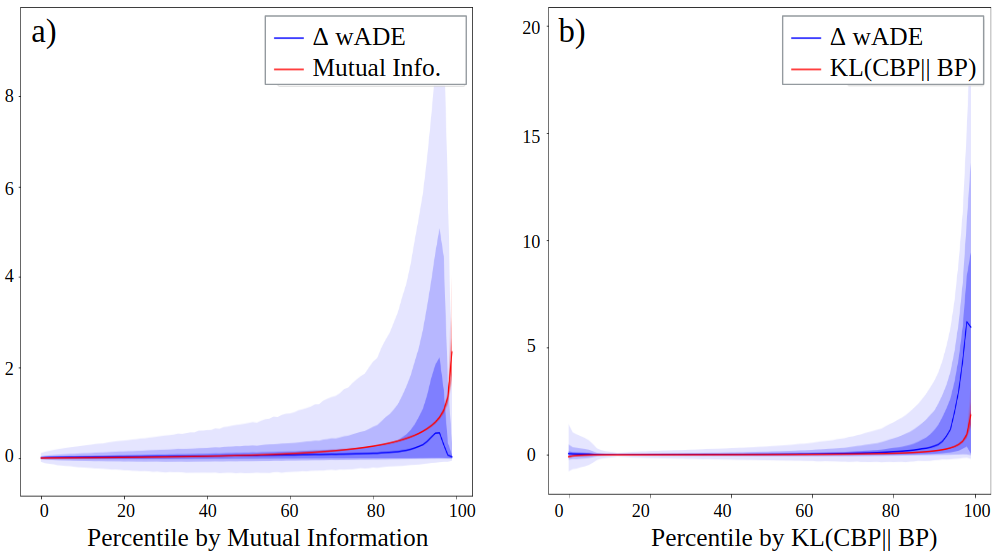}
        \caption{(a) The mutual information between the target and query agents and (b) the KL divergence for the target agent given the ground-truth query trajectory both correlate with the incidence of surprising interactions, as measured by the target's $\Delta_{\text{wADE}}(B)$. Shaded regions are between the 10 \& 90th, 20 \& 80th, 30 \& 70th, and 40 \& 60th percentiles. } \label{delta}
        \vspace{-0.5cm}
  \end{figure}

Figure \ref{delta}a shows that there is a strong correlation between high values of mutual information and high values of $\Delta_{\mbox{wADE}}$.  In other words, agents with high interactivity scores are more likely to exert influence on one another.  Note that the interactivity score does not use any future information, while $\Delta_{\mbox{wADE}}$ does.

Also, in percentiles with high mutual information, there is a high occurrence of examples where the conditional prediction errors are much lower than marginal prediction errors.  These are scenes where the behavior of the query agent has significantly affected the target agent.  Such examples are not present in the lower mutual information percentiles.

On the other hand, Figure \ref{delta}b shows a decrease in average $\Delta_{\text{wADE}}$ for the percentiles with the highest mutual information.  Upon inspection of a portion of scenes in the top percentiles, we observe many examples where the agent pair are positioned very close to each other and can exert influence on one another, however since they are almost stationary and $\Delta_{\text{wADE}}$ is sensitive to distance, the impact of influence on $\Delta_{\text{wADE}}$ is small.
We also observe that a high KL divergence for the target agent given the ground-truth query trajectory strongly correlates with high values of $\Delta_{\mbox{wADE}}$. Given the future trajectory of the query agent, we can predict surprising interactions even more accurately than without future trajectories for either agent.

\begin{figure}[h]
        \centering
    \begin{subfigure}[t]{.49\textwidth}
    \centering
      \includegraphics[width=1.0\textwidth]{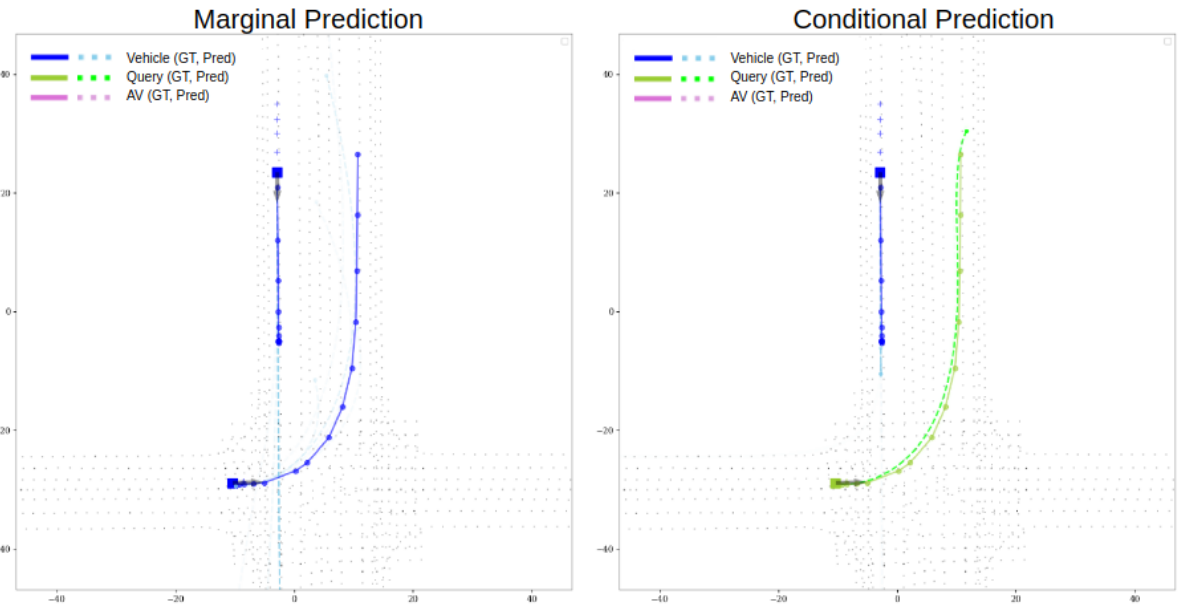}
      \caption{Query agent turns left and target agent yields.}
      \label{example1}
  \end{subfigure}
  \hfill
      \begin{subfigure}[t]{.49\textwidth}
      \centering
      \includegraphics[width=1.0\textwidth]{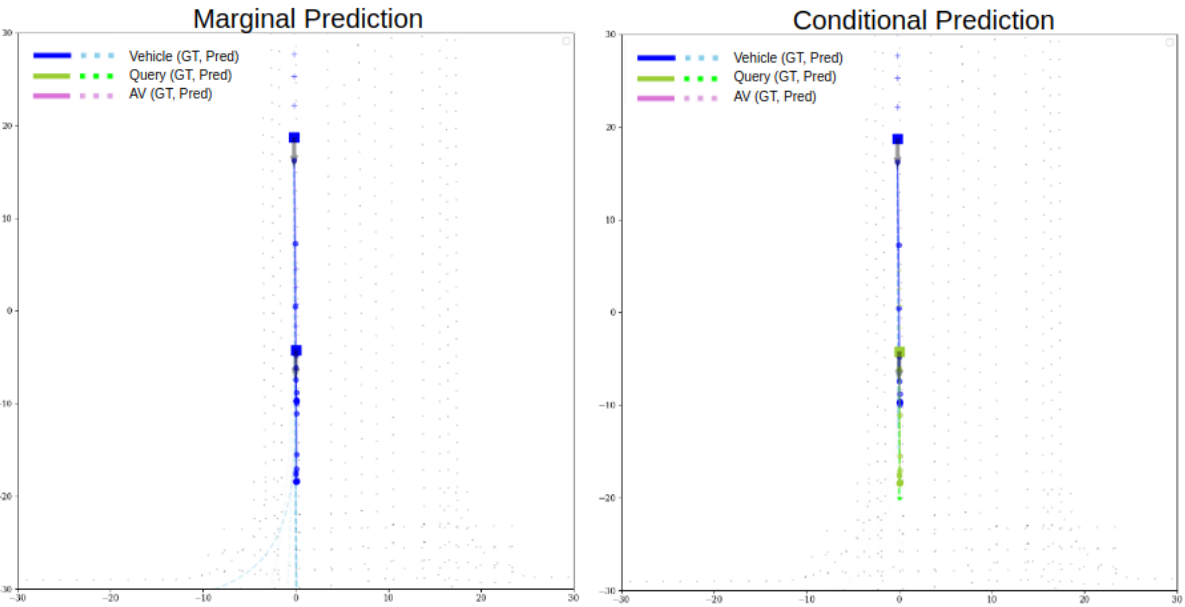}
      \caption{Target agent slows down behind the query agent.}
      \label{example2}
  \end{subfigure}
\caption{Two examples of interacting agents found by sorting examples by mutual information and $\Delta_{\text{wADE}}$. The marginal (left) and conditional predictions (right) are shown with the query in solid green, and predictions in dashed cyan lines.} \label{four_examples}
      \vspace{-0.4cm}
  \end{figure}

Figure \ref{four_examples} shows two examples of pairs of interacting agents discovered in the evaluation set by filtering by high mutual information and high $\Delta_{\text{wADE}}$. In the first example, one vehicle yields to another in a turn. While in the marginal prediction there is a high probability for the target agent to cross the intersection, the conditional prediction shows the target agent yielding. In second example, the target agent slows down behind a query agent which is braking.


\subsection{Selecting Salient Agents}

This section demonstrates using the interactivity score to predict which vehicles are salient for planning for the autonomous vehicle.  
We predict the trajectory of the AV both in the original scene, and in a modified scene where some agents have been removed.  We show agents with high interactivity with the AV are more likely to affect its behavior, compared to agents that are just closer to it.
In the dataset, we typically have 10 to 32 cars in a scene, but in practice, very few of these cars are actually relevant for planning for the AV, so they could potentially be excluded from high-fidelity behavior predictions on-board the vehicle.

In the first experiment, we compute the mutual information between the autonomous vehicle and every other agent. We choose the top $N$ agents with the largest mutual information values. Then, we remove all others from the scene, and use only the top $N$ agents states to predict the trajectory of the AV.  We compare this approach to selecting the top $N$ agents closest in distance to the AV in the scene. This is a common heuristic used for identifying relevant vehicles in the scene.

\begin{figure}[h]
         \vspace{-0.3cm}
        \centering
    \begin{subfigure}[h]{.45\textwidth}
    \centering
      \includegraphics[width=0.75\textwidth]{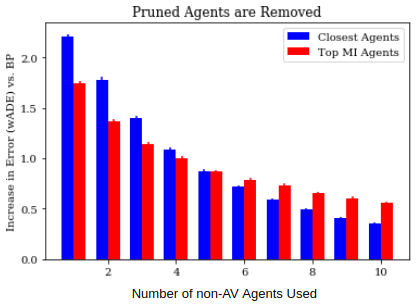}
      \caption{Other agents are removed from the scene.}
      \label{invalid_pruned}
  \end{subfigure}
  \centering
      \begin{subfigure}[h]{.45\textwidth}
      \centering
      \includegraphics[width=0.75\textwidth]{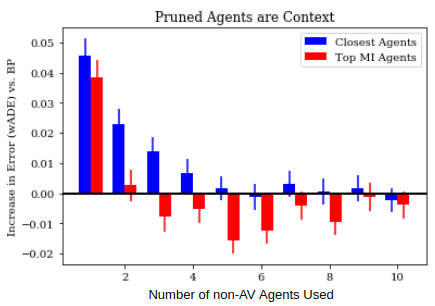}
      \caption{Other agents are used only as context; their behavior is not predicted.}
      \label{context_pruned}
  \end{subfigure}
\caption{The interactivity score allows pruning agents that are not relevant for planning for the AV. The bars show the average BP error over the pruned scene minus the BP error over the original scene. Note that there are no conditional predictions in this experiment.  The error bars indicate the standard error of the mean.} 
\vspace{-0.2cm} 
\label{pruning}
  \end{figure}

Figure \ref{invalid_pruned} shows that mutual information can identify more relevant agents for planning up to $N = 4$.  For larger numbers of agents, the distance heuristic outperforms mutual information as an agent selection mechanism.  In practice, for agent prioritization onboard an AV, mutual information could be combined with other heuristics, such as distance. 


In the second experiment, we do not remove the pruned agents from the scene, but remove them from the set of agents whose behaviors are to be predicted by the model.  In this case, the pruned agents are visible to the model as scene context.  Figure \ref{context_pruned} compares using interactivity score vs. a distance heuristic in this task.  As the results show, moving less interactive agents to scene context is actually improving predictions for the autonomous vehicle, as long as at least the 3 most interactive agents are kept in the prediction set.

One potential explanation for this result is that reducing the prediction set of the model provides an attention mechanism for the prediction of the autonomous vehicle that emphasizes the potential future trajectories of certain agents over others.  In particular, the message-passing mechanism in the GNN can focus only on the relevant neighbors for the AV.

\begin{figure}[h]
      \centering
      \includegraphics[width=0.5\textwidth]{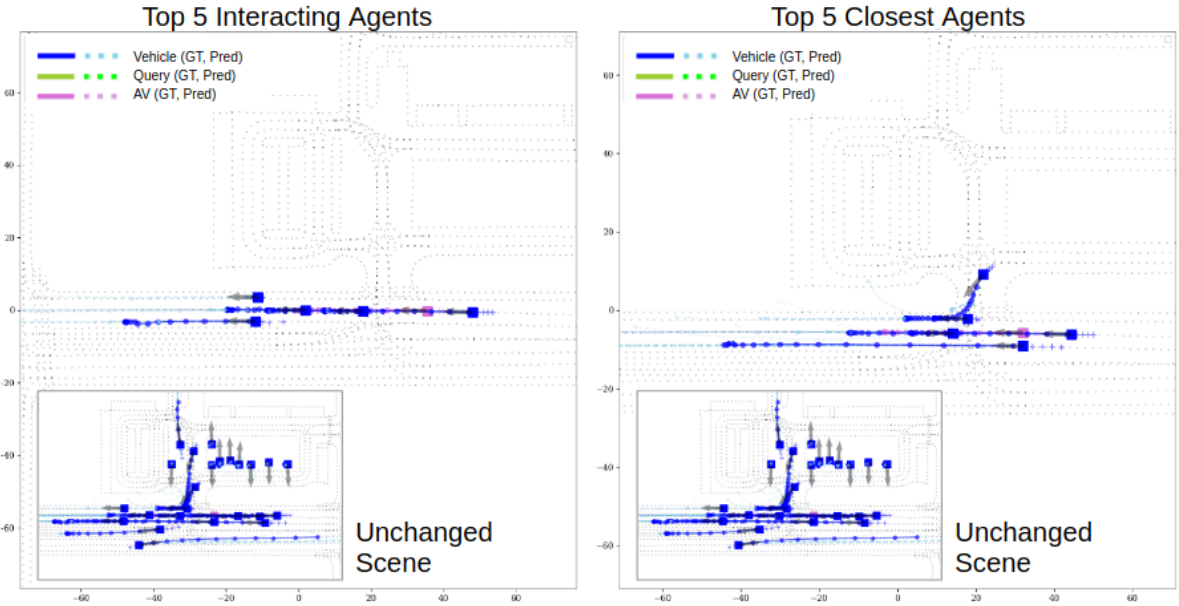}
      \caption{The inset shows the non-pruned scene with the AV (pink) and other cars (blue). On the left, agents with a low interactivity score with the AV are pruned.  On the right, agents are pruned based on distance to the AV. }
\vspace{-0.3cm} 
      \label{example_pruned}
  \end{figure}

Figure \ref{example_pruned} visualizes pruning agents by mutual information vs. pruning by distance in the same scene.  We see that the mutual information selects vehicles that are behind and ahead of the AV in the same lane, in addition to a few vehicles further ahead in neighboring lanes. The distance metric, on the other hand, selects vehicles that are multiple lanes away and are not likely to interact with the AV.

  \begin{figure}[h]
\centering
      \includegraphics[width=0.49\textwidth]{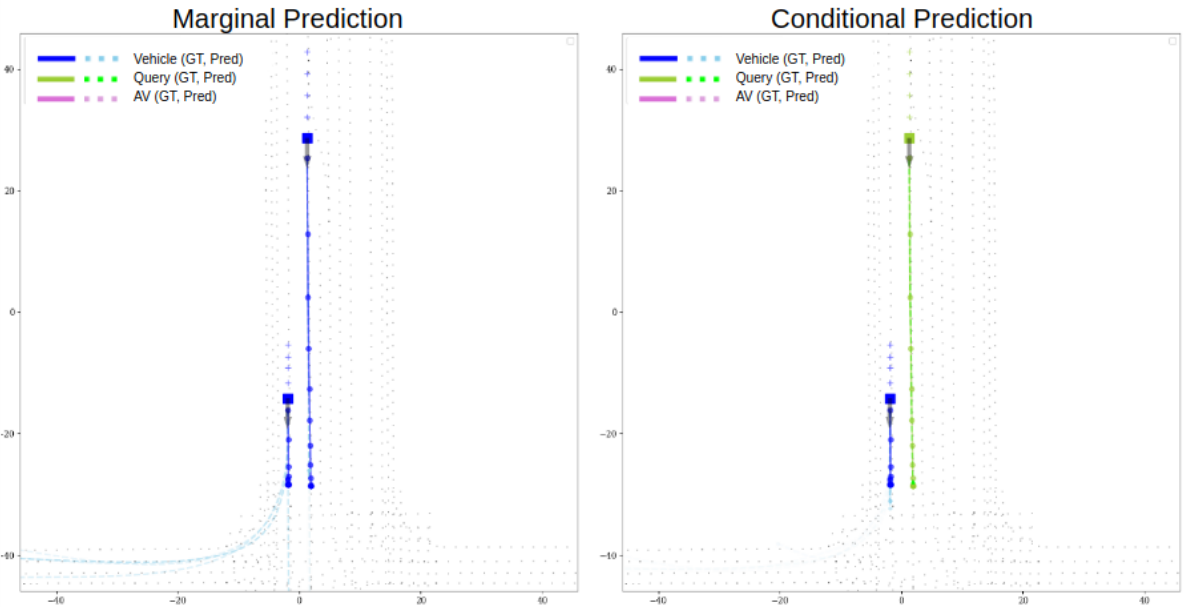}
  \caption{An example in which the query and target agents slow down in parallel lanes as a result of a traffic light change. The marginal (left) and conditional predictions (right) are shown with the query in solid green.         \vspace{-0.5cm}} \label{correlated_examples}
  \end{figure}

\subsection{Challenges and Future Work}

Fig. \ref{correlated_examples} shows an example where our metrics have selected a pair of vehicles slowing down in parallel lanes at an intersection. These agents are reacting to a change in traffic light state, rather than to one another. The CBP model can not differentiate between correlation and causation of two agent's trajectories. 
Before using a trajectory as a query, one can compute the marginal likelihood of the query $p(\bms^B, \bmx)$, to determine whether $\bms^B$ is a likely query for which the model can accurately provide counterfactual predictions. 

The interactivity score can be evaluated very efficiently by pre-computing the embedding of the roadgraph, which is the most expensive part of the architecture in practice, and batching the different queries to evaluate them in parallel.
We could also consider using our interactivity score as a reward signal in cooperative multi-agent reinforcement learning, similar to the notion of influence introduced in~\cite{jaques2019social}.




\clearpage
\newpage
\bibliographystyle{IEEEtran}
\bibliography{bibliography}

\end{document}